\documentclass{article}

\usepackage{microtype}
\usepackage{graphicx}
\usepackage{subfigure}
\usepackage{array, booktabs} 

\usepackage{amsmath,amsfonts,bm}

\def\Figref#1{Figure~\ref{#1}}

\def\eqref#1{equation~\ref{#1}}
\def\Eqref#1{Equation~\ref{#1}}

\def\Algref#1{Algorithm~\ref{#1}}

\def\1{\bm{1}}

\def\vtheta{{\bm{\theta}}}
\def\vepsilon{{\bm{\epsilon}}}

\def\vg{{\bm{g}}}

\def\vx{{\bm{x}}}
\def\vy{{\bm{y}}}

\DeclareMathAlphabet{\mathsfit}{\encodingdefault}{\sfdefault}{m}{sl}
\SetMathAlphabet{\mathsfit}{bold}{\encodingdefault}{\sfdefault}{bx}{n}

\def\gB{{\mathcal{B}}}

\def\gN{{\mathcal{N}}}

\def\gS{{\mathcal{S}}}

\newcommand{\E}{\mathbb{E}}

\newcommand{\Var}{\mathrm{Var}}

\usepackage{pifont}

\usepackage{hyperref}

\usepackage[accepted]{icml2024}

\usepackage{amsmath}
\usepackage{amssymb}
\usepackage{mathtools}
\usepackage{amsthm}
\usepackage{multirow}
\usepackage{nicematrix}

\usepackage[capitalize,noabbrev]{cleveref}

\theoremstyle{plain}
\newtheorem{theorem}{Theorem}[section]

\newtheorem{lemma}[theorem]{Lemma}

\theoremstyle{definition}

\theoremstyle{remark}

\renewcommand{\algorithmiccomment}[1]{//#1}

\usepackage[textsize=tiny]{todonotes}

\icmltitlerunning{When Will Gradient Regularization Be Harmful?}

\begin{document}

\twocolumn[
\icmltitle{When Will Gradient Regularization Be Harmful?}



\icmlsetsymbol{equal}{*}

\begin{icmlauthorlist}
    \icmlauthor{Yang Zhao}{thu}
    \icmlauthor{Hao Zhang}{thu}
    \icmlauthor{Xiuyuan Hu}{thu}
    \end{icmlauthorlist}
    
\icmlaffiliation{thu}{Department of Electronic Engineering, Tsinghua University}
\icmlcorrespondingauthor{Hao Zhang}{haozhang@tsinghua.edu.cn}
\icmlcorrespondingauthor{Yang Zhao}{zhao-yang@tsinghua.edu.cn}

\icmlkeywords{Machine Learning, ICML}

\vskip 0.3in
]



\printAffiliationsAndNotice{}  

\begin{abstract}
    Gradient regularization (GR), which aims to penalize the gradient norm atop the loss function, has shown promising results in training modern over-parameterized deep neural networks. However, can we trust this powerful technique? This paper reveals that GR can cause performance degeneration in adaptive optimization scenarios, particularly with learning rate warmup. Our empirical and theoretical analyses suggest this is due to GR inducing instability and divergence in gradient statistics of adaptive optimizers at the initial training stage. Inspired by the warmup heuristic, we propose three GR warmup strategies, each relaxing the regularization effect to a certain extent during the warmup course to ensure the accurate and stable accumulation of gradients. With experiments on Vision Transformer family, we confirm the three GR warmup strategies can effectively circumvent these issues, thereby largely improving the model performance. Meanwhile, we note that scalable models tend to rely more on the GR warmup, where the performance can be improved by up to 3\% on Cifar10 compared to baseline GR. Code is available at \href{https://github.com/zhaoyang-0204/gnp}{https://github.com/zhaoyang-0204/gnp}.
\end{abstract}

\section{Introduction}

Advancements in computational hardware have catalyzed the design of modern deep neural networks such as the Transformers \cite{DBLP:conf/iclr/DosovitskiyB0WZ21, vaswani2017attention, brown2020language, DBLP:conf/iccv/LiuL00W0LG21, DBLP:conf/iccv/TouvronCSSJ21}, characterized by an extraordinarily vast number of parameters, far exceeding their predecessors. In this context, regularization techniques emerge as a more pivotal role to resist overfitting in training these over-parameterized networks \cite{DBLP:journals/jmlr/SrivastavaHKSS14, DBLP:conf/icml/IoffeS15, DBLP:journals/corr/BaKH16, DBLP:conf/iclr/ForetKMN21}.

\begin{figure}[t]
    \centering
    \includegraphics[width=0.85\columnwidth]{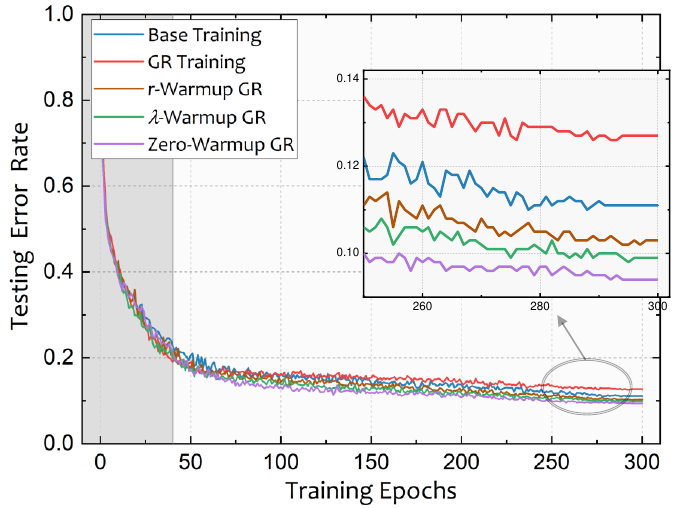}
    \vskip -0.05in
    \caption{Comparison of test error rates (lower values are preferable) of the ViT-B model on the Cifar10 dataset under Base training, GR, and our three proposed GR warmup strategies. All the training instances have also applied the LR warmup. Notably, the performance with LR warmup and normal GR (red line) can be worse compared to training with only LR warmup (blue line).}
    \label{fig : accuracy}
\end{figure}

Recent studies highlight the gradient regularization (GR) as an effective regularization strategy \cite{barrett2020implicit, smith2021origin, zhao2022penalizing, karakida2023understanding, reizinger2023samba}. By imposing an additional penalty concerning gradient norm atop the loss function, this technique deliberately biases the optimization process towards the attainment of flat minima, fostering better generalization. Meanwhile, it has been revealed that a strong association exists between GR and Sharpness-Aware Minimization (SAM) family \cite{DBLP:conf/iclr/ForetKMN21}, which posits SAM as a special parameter configuration in the first-order solution of GR \cite{zhao2022penalizing, karakida2023understanding}.

However, despite its practical utility, the scope and limitations of GR are yet to be fully understood, particularly in terms of establishing when it can be beneficial or safe to apply this technique. We find that GR can lead to serious performance degeneration (see \Figref{fig : accuracy}) in the specific scenarios of adaptive optimization such as Adam \cite{DBLP:journals/corr/KingmaB14} and RMSProp \cite{hinton2012neural}. With both our empirical observations and theoretical analysis, we find that the biased estimation introduced in GR can induce the instability and divergence in gradient statistics of adaptive optimizers at the initial stage of training, especially with a learning rate warmup technique which originally aims to benefit gradient statistics \cite{DBLP:conf/nips/VaswaniSPUJGKP17, DBLP:journals/pbml/PopelB18, DBLP:conf/iclr/LiuJHCLG020}. Notably, this issue tends to become more severe as the complexity of the model increases.

To mitigate this issue, we draw inspirations from the idea of warmup techniques, and propose three GR warmup strategies: $\lambda$-warmup, $r$-warmup and zero-warmup GR. Each of the three strategies can relax the GR effect during warmup course in certain ways to ensure the accuracy of gradient statistics. Then, training reverts to normal GR for the rest of training, allowing the optimization to fully enjoy the performance gain derived from the GR. Finally, we empirically confirm that all the three GR warmup strategies can not only successfully avoid this issue but further enhance the performance for almost all training cases. Of these strategies, the zero-warmup GR can give the best improvements, significant outperforming the baseline.

\section{Background}

\subsection{Gradient Regularization: an Overview}

Gradient regularization typically aims to impose an additional gradient norm penalty on top of the loss function \cite{barrett2020implicit, smith2021origin, zhao2022penalizing, karakida2023understanding},
\begin{equation}
    L^{(gr)}(\vtheta) =  L(\vtheta) + \lambda ||\nabla_{\vtheta} L(\vtheta)||_2
\end{equation}
where $\vtheta$ is the model parameter and $\lambda$ denotes the regularization degree, effectively controlling the extent to which this regularization influences the overall process. Note that some research choose to penalize the $||\nabla_{\vtheta} L(\vtheta)||_2^2$, which actually results in the same effect as $||\nabla_{\vtheta} L(\vtheta)||_2$.

In practical applications, Hessian-free techniques are employed to approximate the involved gradient $L^{(gr)}(\vtheta)$,
\begin{equation}
    \label{eqn : gradient norm reg}
    \begin{split}
        g^{(gr)} = (1 - \frac{\lambda}{r}) \nabla_{\vtheta} L(\vtheta) + \frac{\lambda}{r} \nabla_{\vtheta} L(\vtheta + \vepsilon)
    \end{split}
\end{equation}
with
\begin{equation}
    \begin{split}
        \vepsilon = r \cdot \frac{\nabla_{\vtheta} L(\vtheta)}{|| \nabla_{\vtheta} L(\vtheta) ||_2}
    \end{split}
\end{equation}
where the parameter $r$ denotes as a small, positive scalar representing neighborhood perturbation, which is used for Taylor approximation. 

Notably, the gradients in SAM \cite{DBLP:conf/iclr/ForetKMN21}  can be expressed as,
\begin{equation}
    \label{eqn : gradient norm sam}
    \begin{split}
        g^{(sam)} = \nabla_{\vtheta} L(\vtheta + \vepsilon)
    \end{split}
\end{equation}
which clearly means that SAM essentially regularizes gradient norm with a special parameter configuration in GR where $\lambda = r$ for any case.

\subsection{Related Works}

Recent research has been burgeoning in the field of GR with a specific emphasis on its applications and effects in deep learning. \citet{barrett2020implicit} along with \citet{smith2021origin} have conducted studies first exploring the aspects of GR in deep learning. They have revealed that the stepped update of standard gradient descent implicitly regularizes the gradient norm through its inherent dynamics, especially when correlated with its continuous equivalent, which is identified as implicit GR. When considering explicit GR, they suggest that applying GR explicitly can yield even better results. Recently, \citet{zhao2022penalizing} and \citet{karakida2023understanding} proposes to directly penalize the gradient norm atop the loss and shows that such techniques can potentially give state-of-the-art performance. Meanwhile, it has been highlighted that GR exhibits a strong correlation with SAM \cite{zhao2022penalizing, karakida2023understanding, reizinger2023samba}.

Considering the close connections between GR and SAM family, we would also like to discuss works associated with flat minima. In \cite{DBLP:journals/neco/HochreiterS97a}, the authors are the first to point out that the flatness of minima could be associated with the model generalization, where models with better generalization should converge to flat minima. And such claim has been supported extensively by both empirical evidences and theoretical demonstrations \cite{DBLP:conf/iclr/KeskarMNST17,DBLP:conf/icml/DinhPBB17}. In the meantime, researchers are also fascinating by how to implement practical algorithms to force the models to converge to such flat minima. By summarizing this problem to a specific minimax optimization, \cite{DBLP:conf/iclr/ForetKMN21} introduce the SAM training scheme, which successfully guides optimizers to converge to flat minima. Since then, many SAM variants are proposed, such as improved approaches \cite{DBLP:conf/icml/KwonKPC21, DBLP:conf/iclr/ZhuangGYCADtD022} and efficient approaches \cite{DBLP:journals/corr/abs-2110-03141, DBLP:journals/corr/abs-2203-09962}, in expectation to contributing the original SAM from various perspectives.

\renewcommand{\arraystretch}{1.1}
\begin{table*}[t]
    \caption{Testing error rate of ViT models using Adam and RMSProp optimizers on Cifar-\{10, 100\} when employing a cross-utilization of LR warmup and standard GR techniques.}
    \begin{center}
    \begin{small}
    \begin{sc}
    \begin{tabular}{lcccccc}
        \toprule
       \multirow{2}{2.4cm}[-0.25em]{Model Architecture}  &  \multirow{2}{1.7cm}[-0.25em]{\centering Base Optimizer} & \multirow{2}{1.7cm}[-0.25em]{\centering LR Warmup} & \multicolumn{2}{c}{~~~Cifar-10 Error [\%]~~~} & \multicolumn{2}{c}{~~~Cifar-100 Error [\%]~~~}
          \\
        \cmidrule{4-7}
         &  &  & Vanilla & GR & Vanilla & GR \\
        \midrule
        \multirow{4}{*}{ViT-Ti} & \multirow{2}{*}{Adam} & \ding{53} & $\ \ 15.94_{ \pm 0.33}\ $ & $\ \ 14.37_{ \pm 0.31}\ $ & $\ \ 41.19_{ \pm 0.21}\ $ & $\ \ 39.99_{ \pm 0.39}\ $ \\
         & & \checkmark & $\ \ 14.82_{ \pm 0.64}\ $ & $\ \ 13.92_{ \pm 0.19}\ $ & $\ \ 39.25_{ \pm 0.41}\ $ & $\ \ 39.28_{ \pm 0.50}\ $ \\
        \cmidrule{2-7}
         & \multirow{2}{*}{RMSProp} & \ding{53} & $\ \ 17.05_{ \pm 0.45}\ $ & $\ \ 16.32_{ \pm 0.30}\ $ & $\ \ 41.62_{ \pm 0.16}\ $ & $\ \ 41.06_{ \pm 0.90}\ $ \\
         & & \checkmark & $\ \ 16.47_{ \pm 0.39}\ $ & $\ \ 15.79_{ \pm 0.25}\ $ & $\ \ 40.31_{ \pm 0.50}\ $ & $\ \ 40.02_{ \pm 0.22}\ $ \\
        \midrule
        \multirow{4}{*}{ViT-S} & \multirow{2}{*}{Adam} & \ding{53} & $\ \ 15.43_{ \pm 0.52}\ $ & $\ \ 14.41_{ \pm 0.42}\ $ & $\ \ 42.83_{ \pm 0.34}\ $ & $\ \ 39.16_{ \pm 0.27}\ $ \\
         & & \checkmark & $\ \ 12.07_{ \pm 0.53}\ $ & \textcolor{red}{$\ \ \mathbf{12.40_{ \pm 0.17}}\ $} & $\ \ 37.17_{ \pm 0.24}\ $ & $\ \ 36.65_{ \pm 0.23}\ $ \\
        \cmidrule{2-7}
         & \multirow{2}{*}{RMSProp} & \ding{53} & $\ \ 19.94_{ \pm 0.44}\ $ & $\ \ 17.74_{ \pm 0.23}\ $ & $\ \ 43.09_{ \pm 0.34}\ $ & $\ \ 40.63_{ \pm 0.30}\ $ \\
         & & \checkmark & $\ \ 15.47_{ \pm 0.15}\ $ & $\ \ 14.77_{ \pm 0.35}\ $ & $\ \ 38.55_{ \pm 0.22}\ $ & $\ \ 37.90_{ \pm 0.19}\ $ \\
         \midrule
         \multirow{4}{*}{ViT-B} & \multirow{2}{*}{Adam} & \ding{53} & $\ \ 18.72_{ \pm 0.21}\ $ & $\ \ 16.11_{ \pm 0.43}\ $ & $\ \ 44.16_{ \pm 0.50}\ $ & $\ \ 41.69_{ \pm 0.42}\ $ \\
          & & \checkmark & $\ \ 10.83_{ \pm 0.28}\ $ & \textcolor{red}{$\ \ \mathbf{12.36_{ \pm 0.33}}\ $} & $\ \ 36.70_{ \pm 0.17}\ $ & $\ \ 36.65_{ \pm 0.26}\ $ \\
         \cmidrule{2-7}
          & \multirow{2}{*}{RMSProp} & \ding{53} & $\ \ 20.19_{ \pm 0.17}\ $ & $\ \ 19.06_{ \pm 0.67}\ $ & $\ \ 44.47_{ \pm 0.47}\ $ & $\ \ 42.28_{ \pm 0.65}\ $  \\
          & & \checkmark & $\ \ 15.76_{ \pm 0.25}\ $ & $\ \ 14.97_{ \pm 0.25}\ $ & $\ \ 39.03_{ \pm 0.34}\ $ & $\ \ 39.04_{ \pm 0.35}\ $ \\
        \bottomrule
    \end{tabular}
    \end{sc}
    \end{small}
    \end{center}
    \label{tbl : base res}
\end{table*}


\section{The Incompatibility Between Gradient Regularization and Adaptive Optimization}

In this section, we will explore the effect of employing gradient regularization in the context of adaptive optimization. Firstly, we would present empirical results derived from practical training scenarios.

\subsection{Practical Training with Gradient Regularization in Adaptive Optimization}

\paragraph{Vanilla Adaptive Optimization}
Adaptive optimization is an advanced optimization algorithm in the training of DNNs. By dynamically adjusting learning rates during training, such a technique can largely improve the training efficiency and model performance, especially on complex, non-stationary problems. Unlike fixed learning rate methods, adaptive optimizers such as Adam and RMSProp can expediently navigate the model towards optimal solutions by tuning learning rates per parameter based on estimations of first and second moments of the gradients. For Adam, its gradient momentum $\phi(g; t)$ and adaptive learning rate $\psi(g; t)$ at step $t$ can be expressed as,
\begin{equation}
    \begin{split}
        \phi(g_0, \cdots, g_t ;t) & = \frac{(1 - \beta_1)\sum_{i = 1}^{t} \beta_1^{t - i}g_i}{1 - \beta_1^t} \\
        \psi(g_0, \cdots, g_t ;t) & = \sqrt{\frac{1 - \beta_2^t}{(1 - \beta_2) \sum_{i = 1}^{t} \beta_2^{t - i}g_i^2}}
    \end{split}
\end{equation}
where $g_t$ represents the gradient at step $t$.

Nonetheless, the initial stages of training often involve inaccurate estimation of gradients. Compared to other optimizers, the intrinsic reliance of adaptive optimizers on gradient moments makes it more difficult to get rid of these misleading estimations later in the process. As a consequence, this can potentially steer the optimization towards converging to suboptimal local minima.

A commonly adopted strategy to mitigate this issue is the implementation of learning rate (LR) warmup techniques. With this method, a lower learning rate than the base rate is applied during the initial stages of training, before being increased to the standard learning rate for the rest of the training period. Specifically, there are two typical LR warmup implementations, constant LR warmup and gradual LR warmup. For constant LR warmup, a low constant learning rate is maintained throughout the warmup period. Whereas for gradual LR warmup, the learning rate will ramp up progressively from a very low value to the normal level, commonly in a linear manner,
\begin{equation}
    \eta(t) = \min \{\frac{t}{T_w}, 1\} \eta_0 
\end{equation}
where $T_w$ denotes the warmup steps and $\eta$ denotes the base learning rate.

\begin{figure*}[ht]
    \centering
    \includegraphics[width=1.8\columnwidth]{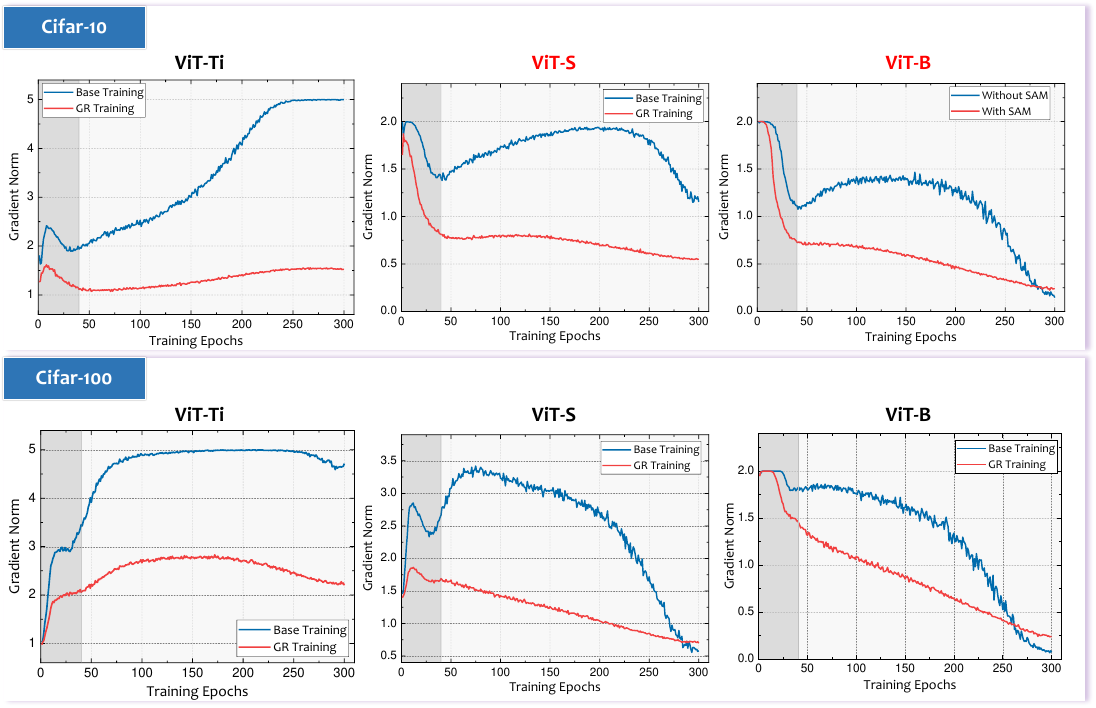}
    \caption{Evolution of gradient norm during training for ViT-Ti, ViT-S and ViT-B models on Cifar-\{10, 100\}, comparing training with GR (red line) and without GR (blue line).}
    \label{fig : gradient norm results}
\end{figure*}

In practice, the gradual LR warmup method is often preferred over its constant warmup. Due to its smoother transition in learning rates, this method often yields better model performance. Accordingly, we will adopt the gradual LR warmup methodology in our subsequent investigations.

\paragraph{Training Implementation}

Firstly, we aim to empirically present the effect of cross-utilizing LR warmup and gradient regularization techniques in the context of adaptive optimization, with a particular emphasis on checking their compatibility in the training process. As for adaptive optimizers, we will choose the Adam optimizer and RMSProp optimizers. 

For the network models, we will focus on the prevalent Vision Transformer \cite{DBLP:conf/iclr/DosovitskiyB0WZ21} models, considering that the conventional CNN models are commonly trained via the SGD optimizer to give better performance. We select to train ViT-Ti, ViT-S and ViT-B model architectures from scratch on Cifar-\{10, 100\} dataset. Notably, the model complexity gradually increases from ViT-Ti to ViT-B.

To ensure optimal model performance, we will leverage established training recipes recommended in the contemporary works \cite{DBLP:conf/iclr/DosovitskiyB0WZ21,DBLP:conf/iclr/ForetKMN21,zhao2022penalizing,karakida2023understanding} and additionally conduct searches on essential hyperparameters. Details on the training process can be found in the Appendix. Table \ref{tbl : base res} shows the final training results, reporting both the mean and standard deviation of testing error rates under five different run seeds.

\subsection{Performance Degradation When Using GR and Adaptive Optimization Simultaneously}

\paragraph{LR warmup can improve training performance.}

From the table, we can clearly see that for all the training  comparison pairs in vertical (training instances with and without the implementation of a LR warmup technique), training with LR warmup will universally give better model performance for both vanilla training and gradient regularization training. Concretely, for both Adam and RMSProp optimizers, leveraging a LR warmup technique generally reduce the testing error rate on Cifar10 and Cifar100 datasets. Notably, in the absence of the LR warmup technique, the performance of ViT-B models, which possess the highest model capacity among the three architectures, degenerates to a catastrophic level, significantly inferior to even the simplest ViT-Ti models. However, with LR warmup, ViT-B models can outperform the other two architectures. The observed enhancement in model performance progressively intensifies from ViT-Ti to ViT-B architectures, suggesting that more complicated models are considerably dependent on the LR warmup technique. Therefore, the LR warmup technique emerges as a indispensable component for adaptive learning algorithms.

Additionally, we can also find that for these ViT models, training with Adam optimizers noticeably excels over RMSProp optimizers, particularly when LR warmup techniques are adopted for more complex models. For instance, when applied to ViT-B models on Cifar10 in vanilla training, the testing error rate when utilizing the Adam optimizer can be approximately 5\% lower compared to the usage of the RMSProp optimizer. Meanwhile, from the improvements, Adam optimizers derive greater benefits from the LR warmup techniques as compared to RMSProp optimizers, especially in training without gradient regularization.

\paragraph{Gradient regularization can have side-effect with LR warmup techniques.} The horizontal training paris in the table (with and without the implementation of GR) elucidates a nuanced relationship between GR and LR warmup techniques. When training without LR warmup, GR consistently enhances model performance across all instances, just as its effectiveness observed in CNN training. However, when GR is combined with LR warmup, GR can be less helpful, as some performances can be far from behind those achieved via vanilla training with LR warmup techniques.

When GR training with LR warmup, for the RMSProp optimizer, GR can marginally improve model performance for most instances. Nevertheless, RMSProp fails to match the competitive performance demonstrated by the Adam optimizer. For the latter, a noticeable degradation in model performance can be observed for certain training instances. In these cases, models trained with both GR and LR warmup yield higher testing error rates compared to their counterparts trained with LR warmup but without GR. Notably, this degeneration becomes more significant as model complexity increases. This suggests that GR and warmup can be incompatible when employed simultaneously, especially for large models. Coupled with our previous observation that larger models rely more on the LR warmup technique, we can reason that larger architectural models like ViT-B are more sensitive to gradient statistics at the inception of the training phase.

\section{Understanding the Incompatibility Between GR and Adaptive Optimization}

In this section, we delve deeper into the conditions under which this compatibility issue between GR and warmup techniques occur, and perform a theoretical analysis to understand their underlying causes.

\paragraph{Compatibility issues arise with an initial decline in gradient norm.} Recall that GR essentially regularizes the norm of loss gradient with a degree $\lambda$. It implies that once training starts, GR will bias the gradient towards regions where the norm is smaller. Such biased estimation may potentially undermine the stable and precise accumulation of gradient statistics at the initial stage of training. Consequently, this motivates us to check the evolution of the gradient norm during training. \Figref{fig : gradient norm results} shows this evolution across epochs when employing the Adam optimizer for training with a LR warmup duration of 40 epochs, where the gradient norm is averaged over each individual step within a single epoch.

From the figure, we can clearly see that the gradient norm while training with GR is significantly smaller than that observed without GR across all training instances. Combined with previous incompatible training instances in Table \ref{tbl : base res}, we can find that these instances all originally exhibit a decrease in gradient norm at the initial stages of training and the use of GR further exacerbates this decline. In other words, if the original gradient norm experiences a decline, then imposing gradient regularization can trigger compatibility issues.

\paragraph{GR can lead to large adaptive leraning rate variance in theory}

Next, we will give theoretical analysis regarding this compatibility from the perspective of adaptive LR variance.

Consider gradients ${g_1, \cdots, g_t}$, where each element $g_k$ is independent and conforms $g_k \sim \gN (0, \sigma_k)$. We target to investigate the variance of the adaptive LR $\psi(\cdot)$, in the case where gradient norm decreases as $t$ increases. It is important to note that this scenario differs from the one proposed in \cite{DBLP:conf/iclr/LiuJHCLG020}, where $g_k \sim \gN (0, \sigma)$ consistently.

A decrease in gradient norm generally indicates a contraction in the overall magnitude of the gradients. Given that these gradients are from a Gaussian distribution $g_k \sim \gN (0, \sigma_i)$, a decreased gradient norm implies a corresponding drop in variance since variance quantifies the extent of deviation around the mean. Thereby, we assume the variance $\sigma_k$ decays exponentially,
\begin{equation}
    \sigma_k = e^{-\gamma k}\sigma_1
\end{equation}

Now, the distribution of $\psi(\cdot)$ follows a nonstandard and intricate distribution. Perform a straightforward analysis can be challenging. Similar to that in \cite{DBLP:conf/iclr/LiuJHCLG020}, we adopt the approximation proposed by \cite{nau2014forecasting}, which is $p(\psi(\cdot;\gamma)) \approx p(\sqrt{\frac{t}{\sum_{k = 1}^t g_k^2}})$.

\begin{lemma}\label{lemma1}
    The variance of the adaptive learning rate can be evaluated via the Taylor series method \cite{benaroya2005probability},
    \begin{equation}\label{eqn : variance approx}
        \Var(\psi(\cdot;\gamma)) \approx \frac{\Var(Y)}{4\E(Y)^3}
    \end{equation}
    where $Y = \frac{1}{t} \sum_{k = 1}^t g_k^2$
\end{lemma}

By leveraging the Taylor approximation, Lemma \ref{lemma1} provides a valid way for calculating the adaptive LR variance. The proof can be found in the Appendix. 

\begin{figure}[t]
    \centering
    \includegraphics[width=0.9\columnwidth]{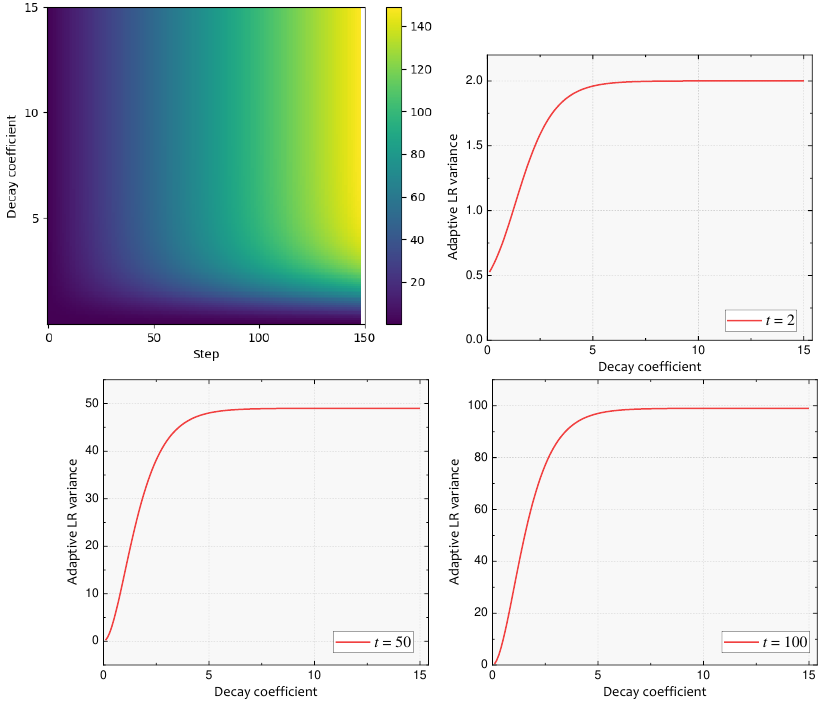}
    \vskip -0.05in
    \caption{Plot of the \Eqref{eqn : variance computation}}
    \label{fig : varaince}
\end{figure}

\begin{theorem}
    Given $g_k$ are independent, and each follows $g_k \sim \gN(0, \sigma_k)$, where $\sigma_k = e^{-\gamma k}\sigma_1$, $\gamma > 0$. $\Var(\psi(\cdot;\gamma))$ is given by,
    \begin{equation}\label{eqn : variance computation}
        \Var(\psi(\cdot;\gamma)) = \frac{t}{2\sigma_1} \frac{(1 + e^{-\gamma t})(1 - e^{-\gamma})^2}{(1 + e^{-\gamma})(1 - e^{-\gamma t})^2}
    \end{equation} 
\end{theorem}

Here, we provide a plot for the \Eqref{eqn : variance computation}, as shown in \Figref{fig : varaince}. In the figure, $x$-axis denotes the step $t$ while $y$-axis denotes the decay coefficient $\gamma$. We can clearly see that the adaptive LR variance $\Var(\psi(\cdot;\gamma))$ increases along with the decay coefficient $\gamma$.

\begin{theorem}\label{thm : varaince proof}
    For $t e^{-\gamma (t - 1)} \leq 1$, $\Var(\psi(\cdot;\gamma))$ monotonically increases as $\gamma$ increases.
\end{theorem}

Also, the proof can be found in the Appendix. And actually, $t e^{-\gamma (t - 1)} \leq 1$ would hold almost, as $t$ keeps increasing.

Theorem \ref{thm : varaince proof} theoretically confirms that $\Var(\psi(\cdot;\gamma))$ increases as $\gamma$ increase. In conclusion, this indicates that GR will lead to a larger adaptive LR variance as compared to vanilla training in scenarios involving a decline in gradient norm. This is the fundamental reason for the observed incompatibility between GR and adaptive optimization techniques.

\section{GR Warmup Strategies}

\begin{figure*}[ht]
    \centering
    \includegraphics[width=1.8\columnwidth]{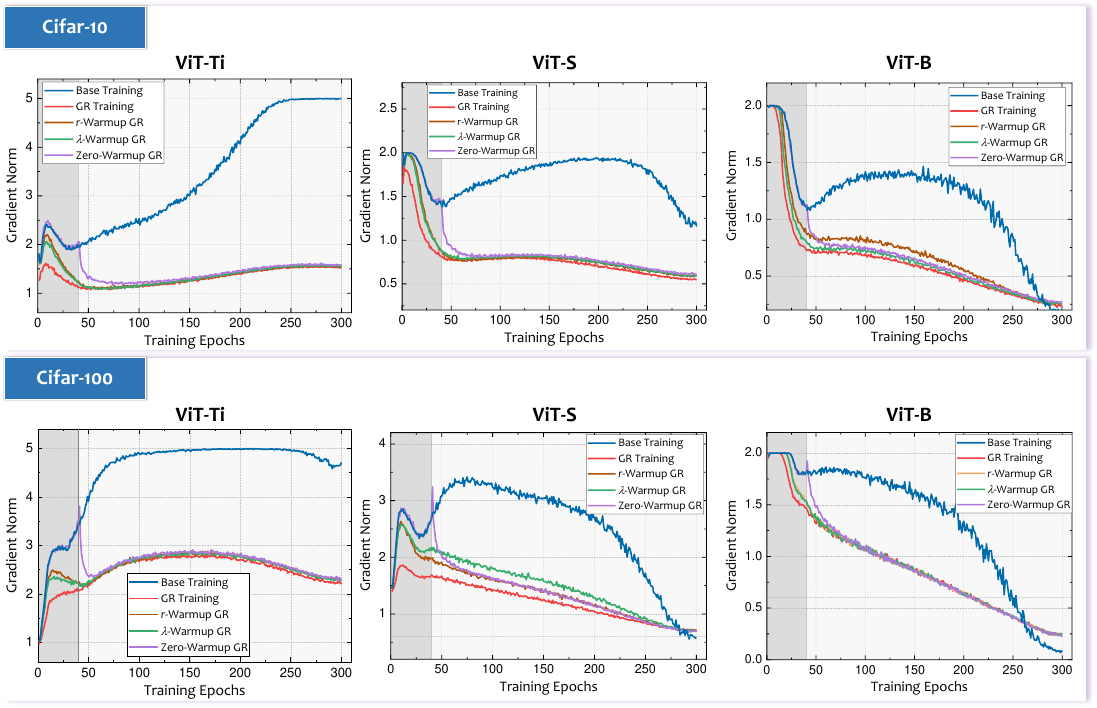}
    \vskip -0.05in
    \caption{Evolution of gradient norm during training for ViT-Ti, ViT-S and ViT-B models on Cifar-\{10, 100\}, comparing training with $r$-warmup GR strategy (dark brown line), $\lambda$-Warmup GR strategy (green line) and zero-warmup GR strategy (purple line).}
    \label{fig : gradient norm results for warmup strategy}
\end{figure*}

As indicated by previous demonstrations, the current GR will impose excessive regularization on gradient norm during warmup course. This results in a substantial increase in adaptive LR variance, especially when the gradient norm originally experiences a drop. To mitigate this issue, it is motivated to reduce the regularization degree on gradient norm at the beginning of training, analogous to the LR warmup techniques.

Here, we will introduce three GR warmup strategies: $r$-warmup, $\lambda$-warmup and zero-warmup. \Algref{alg:algorithm} shows the general framework of training with our warmup strategies. Table \ref{tbl : advanced warmup} shows the summary of the model performance and \Figref{fig : gradient norm results for warmup strategy} shows the corresponding evolution of gradient norm when applying the three warmup strategies. 

All the training details can be found in the Appendix. Additionally, we have included further results on various model architectures and datasets in the Appendix.

\begin{algorithm}[tb]
    \caption{Gradient Regularization Warmup Strategies}
    \label{alg:algorithm}
    \textbf{Input}: Training set $\gS = \{(\vx_i, \vy_i)\}_{i = 0}^{N}$; loss function $L(\cdot)$; batch size $B$; learning rate $\eta_0$; total iterations $T$; warmup iterations $T_w$; GR coefficient $\lambda_0$ and $r_0$; GR warmup strategies $\mathcal{P}$. \\
    \textbf{Parameter}: Model weights $\vtheta$. \\
    \textbf{Output}: Optimized model weights $\hat{\vtheta}$. \\
    \textbf{Algorithm}:
    \begin{algorithmic}[1]
    \STATE {Parameter initialization $\vtheta_{0}$.}
    \FOR{step $t=1$ {\bfseries to} $T$}
    \STATE{Get batch data pairs $\gB = \{(\vx_i, \vy_i)\}_{i = 0}^{B}$ sampled from training set $\gS$.}
    \IF{$t < T_w$}
    \STATE $\eta(t) = \min \{\frac{t}{T_w}, 1\} \eta_0 $
    \STATE Perform $\mathcal{P} \in $ \{$r$, $\lambda$, zero\}-warmup GR strategies
    \ELSE
    \STATE Training with $\eta_0$, $\lambda_0$ and $r_0$.
    \ENDIF
    \STATE Compute the gradient $\vg^{(1)} = \nabla_\vtheta L(\vtheta_t)$.
    \IF{$r_t > 0$}
    \STATE $\vg^{(2)} = \nabla_\vtheta L(\vtheta_t)$ at ${\vtheta_t = \vtheta_t + r_t \frac{\vg^{(1)}}{||\vg^{(1)}||}}$.
    \ELSE 
    \STATE $\vg^{(2)} = 0$
    \ENDIF
    \STATE $\vg_t = (1 - \frac{\lambda_t}{r_t}) \vg^{(1)} + \frac{\lambda_t}{r_t} \vg^{(2)}$
    \STATE Update weight $\vtheta_{t + 1} = \vtheta_{t} - \eta \cdot \vg_t$
    \ENDFOR
    \STATE{\textbf{return} Final optimization $\hat{\vtheta}$}.
    \end{algorithmic}
\end{algorithm}

\begin{table*}[t]
    \caption{Test error rate of ViT models using Adam and RMSProp optimizers on the Cifar-\{10, 100\} datasets when trained with our three proposed GR warmup strategies}
    \begin{center}
    \begin{small}
    \begin{sc}
    \begin{tabular}{lcccc}
        \toprule
        Model & ~Base~ & ~~~GR Warmup~~~ & ~~~Cifar-10~~~ & ~~~Cifar-100~~~ \\
        Architecture~~~& Optimizer & Policy & Error[\%] & Error[\%] \\
        \midrule
        \multirow{8}{*}{ViT-Ti} & \multirow{4}{*}{Adam} & GR (baseline)  & $\ \ 13.92_{ \pm 0.19}\ $ & $\ \ 39.38_{ \pm 0.50}\ $ \\
        &  & $r$-Warmup  & $\ \ 13.80_{ \pm 0.17}\ $ & $\ \ 39.24_{ \pm 0.37}\ $ \\
        &  & $\lambda$-Warmup  & $\ \ 13.86_{ \pm 0.35}\ $ & $\ \ \mathbf{39.12_{ \pm 0.37}}\ $ \\
        &  & Zero-Warmup  & $\ \ \mathbf{13.61_{ \pm 0.20}}\ $ & $\ \ 38.53_{ \pm 0.38}\ $ \\
        \cmidrule{2-5}
        & \multirow{4}{*}{RMSProp} & GR (baseline) & $\ \ 15.79_{ \pm 0.25}\ $ & $\ \ 40.02_{ \pm 0.22}\ $ \\
        &  & $r$-Warmup  & $\ \ 15.32_{ \pm 0.21}\ $ & $\ \ 39.80_{ \pm 0.22}\ $ \\
        &  & $\lambda$-Warmup  & $\ \ 15.47_{ \pm 0.27}\ $ & $\ \ 39.55_{ \pm 0.19}\ $ \\
        &  & Zero-Warmup  & $\ \ \mathbf{15.06_{ \pm 0.40}}\ $ & $\ \ \mathbf{39.25_{ \pm 0.21}}\ $ \\
        \midrule
        \multirow{8}{*}{ViT-S} & \multirow{4}{*}{Adam} & GR (baseline) & $\ \ 12.40_{ \pm 0.17}\ $ & $\ \ 36.65_{ \pm 0.23}\ $ \\
        &  & $r$-Warmup  & $\ \ 11.72_{ \pm 0.38}\ $ & $\ \ 36.53_{ \pm 0.18}\ $ \\
        &  & $\lambda$-Warmup  & $\ \ 11.47_{ \pm 0.04}\ $ & $\ \ 36.48_{ \pm 0.23}\ $ \\
        &  & Zero-Warmup  & $\ \ \mathbf{10.68_{ \pm 0.49}}\ $ & $\ \ \mathbf{35.41_{ \pm 0.25}}\ $ \\
        \cmidrule{2-5}
        & \multirow{4}{*}{RMSProp} & GR (baseline) & $\ \ 14.77_{ \pm 0.35}\ $ & $\ \ 37.90_{ \pm 0.19}\ $ \\
        &  & $r$-Warmup  & $\ \ 14.73_{ \pm 0.18}\ $ & $\ \ 37.65_{ \pm 0.30}\ $ \\
        &  & $\lambda$-Warmup  & $\ \ 14.65_{ \pm 0.39}\ $ & $\ \ 37.74_{ \pm 0.27}\ $ \\
        &  & Zero-Warmup  & $\ \ \mathbf{14.02_{ \pm 0.29}}\ $ & $\ \ \mathbf{37.37_{ \pm 0.31}}\ $ \\
        \midrule
        \multirow{8}{*}{ViT-B} & \multirow{4}{*}{Adam} & GR (baseline) & $\ \ 12.36_{ \pm 0.33}\ $ & $\ \ 36.65_{ \pm 0.26}\ $ \\
        &  & $r$-Warmup  & $\ \ 10.22_{ \pm 0.26}\ $ & $\ \ 36.18_{ \pm 0.37}\ $ \\
        &  & $\lambda$-Warmup  & $\ \ 9.98_{ \pm 0.23}\ $ & $\ \ 36.03_{ \pm 0.30}\ $ \\
        &  & Zero-Warmup  & $\ \ \mathbf{9.42_{ \pm 0.15}}\ $ & $\ \ \mathbf{35.75_{ \pm 0.20}}\ $ \\
        \cmidrule{2-5}
        & \multirow{4}{*}{RMSProp} & GR (baseline) & $\ \ 14.97_{ \pm 0.25}\ $ & $\ \ 39.04_{ \pm 0.35}\ $ \\
        &  & $r$-Warmup  & $\ \ 14.42_{ \pm 0.11}\ $ & $\ \ 38.36_{ \pm 0.22}\ $ \\
        &  & $\lambda$-Warmup  & $\ \ 14.55_{ \pm 0.39}\ $ & $\ \ 38.50_{ \pm 0.15}\ $ \\
        &  & Zero-Warmup  & $\ \ \mathbf{14.34_{ \pm 0.24}}\ $ & $\ \ \mathbf{38.24_{ \pm 0.21}}\ $ \\
        \bottomrule
    \end{tabular}
    \label{tbl : advanced warmup}
\end{sc}
\end{small}
\end{center}
\end{table*}

\subsection{$r$-warmup Gradient Regularization}

The $r$-warmup GR is designed with due consideration to not only GR but also SAM. It is important to recall that the regularization degree $r^{(sam)}$ in SAM corresponds to the parameter setting $r = \lambda$ in GR. As such, our first strategy involves concurrently ramping up parameters $r$ and $\lambda$ in GR (which equates to $r^{(sam)}$ in SAM), following the same pattern as that of the learning rate warmup schedule.
\begin{equation}
    r_t =   \min \{\frac{t}{T_w}, 1\} \cdot r_0
\end{equation}
\noindent where $r_0$ denotes the base value we set. Correspondingly, the gradient becomes,
\begin{equation}
    \label{eqn : $r$-Warmup gradient}
    g_t = \nabla_{\vtheta} L(\vtheta_t + r_t \cdot \frac{\nabla_{\vtheta} L(\vtheta_t)}{||\nabla_{\vtheta} L(\vtheta_t)||}) 
\end{equation}
\noindent This approach results in a considerably milder regularization effect initially, which then linearly scales up to its standard strength by the end of the warmup course.

As illustrated in \Figref{fig : gradient norm results for warmup strategy}, the gradient norm is regularized at a slower rate during the initial stages of training, situating the gradient norm between those trained with and without GR. As we can see from Table \ref{tbl : advanced warmup}, when using the $r$-warmup GR during the warmup course, all training cases exhibit improved model performance compared to standard GR implementation. The previously observed degradation effect on model performance can no longer be seen. Particularly for ViT-B models, performance significantly surpasses that of models trained with GR, and slightly outperform those trained without GR. However, although $r$-warmup GR leads to some improvement in model performance, this enhanced performance remains relatively close to that before degradation, i.e., training without GR.

\subsection{$\lambda$-warmup Gradient Regularization}

We note that according to \cite{zhao2022penalizing}, a small $r$ in GR can cause loss of precision since the perturbed gradient term $\nabla_{\vtheta} L(\vtheta_t + r \cdot {\nabla_{\vtheta} L(\vtheta_t)} / {||\nabla_{\vtheta} L(\vtheta_t)||})$ will become exceedingly close to the reference gradient $\nabla_{\vtheta} L(\vtheta_t)$. Therefore, given that $r$ in SAM corresponds to $\lambda = r$ in GR, $r$-warmup GR can suffer from this issue.

To this end, in our second strategy, we will choose to keep the parameter $r$ fixed, and gradually ramp up the regularization degree $\lambda$ in \Eqref{eqn : gradient norm reg} instead, which denotes that
\begin{equation}
    \lambda_t =   \min \{\frac{t}{T_w}, 1\} \cdot \lambda_0
\end{equation}
\noindent where $\lambda_0$ denotes the base regularization degree. Now, the gradient corresponds to,
\begin{equation}
    \label{eqn : lambda-warmup gradient}
    g_t = (1 - \frac{\lambda_t}{r_0}) \nabla_{\vtheta} L(\vtheta_t) + \frac{\lambda_t}{r_0} \nabla_{\vtheta} L(\vtheta_t + r_0 \frac{\nabla_{\vtheta} L(\vtheta_t)}{||\nabla_{\vtheta} L(\vtheta_t)||})
\end{equation}
Note that $\lambda_t < r_0$ for $t < T_w$, suggesting the term $\lambda_t / r_0 < 1$ during the warmup course. Contrary to the first $r$-warmup GR, this approach mitigates the regularization of gradient norm by interpolating the vanilla base (in this case, Adam) gradient and progressively diminishing its ratio in this interpolation. In other words, for $\lambda$-warmup GR, the corresponding gradient starts from the vanilla base gradient at at the beginning of the warmup course and eventually evolve to the normal GR effect at the end of the warmup course.

From the Table \ref{tbl : advanced warmup} and \Figref{fig : gradient norm results for warmup strategy}, we can find that although the gradients between adopting $r$-warmup GR and $\lambda$-warmup GR are quite close, $\lambda$-warmup GR can provide relatively better results. This suggests that performing warmup on $\lambda$ in GR, rather than $r$ in SAM, can effectively mitigate the precision loss problems caused by a small $r$ value. Furthermore, we can find in the figure that while the regularization effects of both the $r$-warmup and $\lambda$-warmup GR are somewhat weaker than the normal GR implementation, they ultimately converge to an equivalent degree of regularization. Consequently, these observations motivate the development of our third strategy.

\subsection{Zero-warmup Gradient Regularization}

Our third strategy will further reduce the gradient norm regularization to a zero effect. In other words, we will use the vanilla base learning algorithm during warmup course, which will not impose any regularization on gradient norm. Subsequently, we switch to the normal GR for the remaining training iterations,
\begin{equation}
    \lambda_t = \begin{cases}
        0   &~~t \leq T_w  \\
        \lambda_0 & ~~ t > T_w        
    \end{cases}
\end{equation}
The corresponding gradient can be expressed as,
\begin{equation}
    g_t = \begin{cases}
        \nabla_{\vtheta} L(\vtheta_t)   &~~t \leq T_w  \\
        g_{t}^{(gr)} & ~~ t > T_w
    \end{cases}
\end{equation}

In this strategy, warmup technique and GR are implemented separately across two distinct training periods. During the first period, we conduct warmup without imposing any regularization, resulting in a gradient norm identical to that obtained when training with the vanilla learning algorithm. This allows the optimizer to accumulate sufficient accurate gradients in statistics by the completion of the warmup phase. After that, the training process enters into the second period, which starts to regularize the gradient norm with normal GR method. Given the accumulation of gradients during the initial period, biasing the gradient towards convergence to regions with smaller gradient norm becomes considerably safer. It is worth noting that since zero-warmup employs base learning instead of sharpness-aware learning, it can further reduce computation overhead to some extent, dependent upon the number of warmup epochs.

As illustrated in the \Figref{fig : gradient norm results for warmup strategy}, as expected, the evolution of gradient norms during the warmup phase when trained with the zero-warmup strategy matches precisely with that in vanilla training. But after that, as the zero-warmup GR switch to normal GR implementation, the gradient norm can be further reduced. For all three strategies, the corresponding gradient norms gradually become quite close, and eventually terminate at nearly the same values. This is because, after the warmup course, training is set to regularize the gradient norm to the same degree in all three warmup GR strategies, regardless of the level of gradient norm regularization applied during the warmup course.

We can see from the Table \ref{tbl : advanced warmup} that the zero-warmup GR yields superior performance compared to the other two strategies, significantly surpassing the benchmark method which involves training with normal GR method. For ViT-B models, this enhancement approximates nearly a 3\% percent notch up on the Cifar10 dataset. These findings highlight that it is best to train with the base optimizer during the warmup course such that the variance of adaptive learning rate can be limited. Once the training stabilizes, we can safely bias the training with GR to achieve better model performance.

\section{Conclusion}

We show that GR and adaptive optimization can be incompatible when applied simultaneously, resulting in performance degeneration. Our empirical and theoretical evidences suggest that this incompatibility is caused by the excessive GR at the initial stage of training. To solve this issue, we propose three GR warmup strategies: $r$-warmup, $\lambda$-warmup, zero-warmup. We confirm the three strategies can successfully avoid this issue and give better performance.

\section*{Impact Statement}

We believe the impacts of our work is positive, can be contributive to training modern complex neural networks. We claim our paper presents no potentially negative implications concerning ethical considerations or future societal consequences.

\section*{Acknowledgements}

We would like to thank all the reviewers and the meta-reviewer for their helpful comments and advices. We would like to thank Yuhan Li and Zhao Shan from Intelligence Sensing Lab at Tsinghua University for the discussions.

\nocite{langley00}

\bibliography{example_paper}
\bibliographystyle{icml2024}

\newpage
\appendix
\onecolumn

\section{Proof of Theorems}

\begin{lemma}
    The variance of the adaptive learning rate can be evaluated via the Taylor series method \cite{benaroya2005probability},
    \begin{equation}
        \Var(\psi(\cdot;\gamma)) \approx \frac{\Var(Y)}{4\E(Y)^3}
    \end{equation}
    where $Y = \frac{1}{t} \sum_{k = 1}^t g_k^2$
\end{lemma}

\emph{Proof:}

We have $\Var(\psi(\cdot;\gamma)) = \sqrt{\frac{1}{Y}}$, where $Y = \frac{1}{t} \sum_{k = 1}^t g_k^2$. Assume $g(X) = \sqrt{\frac{1}{X}}$, where $X$ is a random variable with mean $\mu$. Based on the Taylor series, 
\begin{equation}
    g(X) = g(\mu) + (X - \mu)g^{\prime}(\mu) + \frac{(X - \mu)^2}{2}g{\prime \prime}(\mu) + \cdots
\end{equation}
So, the variance can be,
\begin{equation}
    \begin{split}
        \Var [g(X)] & = \Var\left[g(\mu) + (X - \mu)g^{\prime}(\mu) + \frac{(X - \mu)^2}{2}g{\prime \prime}(\mu)+ \cdots \right]  \\
        & = \Var\left[(X - \mu)g^{\prime} + \frac{(X - \mu)^2}{2}g{\prime \prime}(\mu) + \cdots \right] \\
    \end{split}
\end{equation}

Often the first item is taken,
\begin{equation}
    \Var [g(X)] \approx g^{\prime}(\mu)^2 \Var(X)
\end{equation}

So, considering that $\Var(\psi(\cdot;\gamma)) = g(Y) = \sqrt{\frac{1}{Y}}$, we have,
\begin{equation}
    \begin{split}
        \Var [g(Y)] & \approx g^{\prime}(\mu)^2 \Var(Y) \\
        & = \left(-\frac{1}{2\E(Y)^{\frac{3}{2}}}\right)^2 \Var(Y) \\
        & = \frac{\Var(Y)}{4\E(Y)^3}
    \end{split}
\end{equation}

\begin{theorem}
    Given $g_k$ are independent, and each follows $g_k \sim \gN(0, \sigma_k)$, where $\sigma_k = e^{-\gamma k}\sigma_1$, $\gamma > 0$. $\Var(\psi(\cdot;\gamma))$ is given by,
    \begin{equation}
        \Var(\psi(\cdot;\gamma)) = \frac{t}{2\sigma_1} \frac{(1 + e^{-\gamma t})(1 - e^{-\gamma})^2}{(1 + e^{-\gamma})(1 - e^{-\gamma t})^2}
    \end{equation} 
\end{theorem}

\emph{Proof:}

Assume $\sigma_k = e^{-\gamma k} \sigma_1$, then $\E(g_k^2) = \sigma_k$, $\Var(g_k^2) = 2 \sigma_k^2$. 

So,

\begin{equation}
    \begin{split}
    \E(tY) & = \E(\sum_{k = 1}^t g_{k}^2) \\ 
    & = \sum_{k = 1}^t \E(g_k^2) = \sum_{k = 1}^t \sigma_k = \sum_{k = 1}^t e^{-\gamma} \sigma_1 \\
    & = \sigma_1 \frac{1 - e^{-\gamma t}}{1-e^{-\gamma}}
    \end{split}
\end{equation}

\begin{equation}
    \begin{split}
    \Var(tY) & = \Var(\sum_{k = 1}^t g_{k}^2) \\ 
    & = \sum_{k = 1}^t \Var(g_k^2)  = \sum_{k = 1}^t 2 \sigma_k^2 = \sum_{k = 1}^t 2  (e^{-\gamma} \sigma_1)^2\\
    & = 2\sigma_1^2 \frac{1 - (e^{-\gamma})^{2t}}{1-(e^{-\gamma})^2}
    \\
    & = 2\sigma_1^2 \frac{1 - e^{-2\gamma t}}{1 - e^{-2\gamma}}
\end{split} 
\end{equation}

This lead to,
\begin{equation}
    \begin{split}
        \Var(\psi(\cdot;\gamma)) &  \approx \frac{\Var(Y)}{4\E(Y)^3} \\
        & = \frac{2\frac{1}{t^2}\sigma_1^2 \frac{1 - e^{-2\gamma t}}{1 - e^{-2\gamma}}}{4\left(\frac{1}{t} \sigma_1 \frac{1 - e^{-\gamma t}}{1-e^{-\gamma}} \right)^3} \\
        & = \frac{t}{2\sigma_1} \frac{(1 + e^{-\gamma t})(1 - e^{-\gamma})^2}{(1 + e^{-\gamma})(1 - e^{-\gamma t})^2}
    \end{split}
\end{equation}

\begin{theorem}
    For $t e^{-\gamma (t - 1)} \leq 1$, $\Var(\psi(\cdot;\gamma))$ monotonically increases as $\gamma$ increases.
\end{theorem}

\emph{Proof:}

Assume $x = e^{-\gamma}$. So,
\begin{equation}
    \begin{split}
        \Var(\psi(\cdot;\gamma)) & \approx (h^{\prime}(\E(Y)))^2\Var(Y) \\
        & = \frac{t}{2\sigma_0} \frac{(1 + x^t)(1 - x)^2}{(1 + x)(1 - x^t)^2}
   \end{split} 
\end{equation}

Set 
\begin{equation}
    \begin{split}
        \beta(x) = \log \Var(\psi(\cdot;\gamma)) = & \log t + \log (1 + x^t) + 2 \log (1 - x) - \log(1 + x) - 2\log (1 - x^t)
    \end{split} 
\end{equation}

\begin{equation}
    \begin{split}
        \frac{d\beta(x)}{dx} = & \frac{t x^{t - 1}}{x^t + 1} + \frac{2tx^{t - 1}}{1 - x^t} - \frac{1}{1+ x} - \frac{2}{1 - x} \\
        & = tx^{t - 1}(\frac{1}{1 + x^t} + \frac{2}{1 - x^t}) - (\frac{1}{1 + x} + \frac{2}{1 - x})
    \end{split} 
\end{equation}

Set
\begin{equation}
    \begin{split}
        \zeta(a) = & \frac{1}{1 + a} + \frac{2}{1 - a} \\ 
        \frac{d\zeta(a)}{da} = & \frac{a^2 + 6a + 1}{(1 - a^2)^2} > 0
    \end{split} 
\end{equation}

Given $a = x^t$ decrease with $t$. $\zeta(t)$ decreases. Namely, $\zeta(t) < \zeta(1), t>1$

\begin{equation}
    \begin{split}
        \frac{1}{1 + x^t} + \frac{2}{1 - x^t} < \frac{1}{1 + x} + \frac{2}{1 - x}
    \end{split} 
\end{equation}

Assume $tx^{t - 1} \leq 1$

\begin{equation}
    \begin{split}
        tx^{t - 1}(\frac{1}{1 + x^t} + \frac{2}{1 - x^t}) < \frac{1}{1 + x} + \frac{2}{1 - x}
    \end{split} 
\end{equation}

That is $\frac{d\beta(x)}{dx} < 0 $, $\Var(\psi(\cdot;\gamma)) $ increase as $x$ decrease. Considering $x = e^{-\gamma}$, we can conclude that $\Var(\psi(\cdot;\gamma)) $ increase as $\gamma$ increase.

\section{Implementation Details}

Our training task will utilize ViT models to address image classification problems on the CIFAR dataset. For data argumentation, the data will be subjected to an augmentation process following a standard strategy that has been widely adopted in contemporary works \cite{DBLP:conf/iclr/ForetKMN21}. This approach involves sequentially processing images through a series of operations, starting with four-pixel padding, followed by random flipping and concluding with random cropping.

To get good model performance, we will partially consult the modern ViT training recipes \cite{DBLP:conf/iclr/DosovitskiyB0WZ21}, and meanwhile perform a slight more searching on core hyperparameters.

\begin{table}[ht]
    \caption{The basic hyperparameters for training ViTs.}
    \centering
    \begin{tabular}{lccc}
      \toprule
      & Adam/RMSProp Training & GR Training \\
      \midrule
      Epoch & 300 & 300  \\
      LR/GR Warmup epoch & \{20, 40, 80\} & \{20, 40, 80\}  \\
      Batch size & 256 & 256  \\
      Gradient Clipping & True & True \\
      Basic learning rate & \{5e-3, 1e-3, 5e-4\} & \{5e-3, 1e-3, 5e-4\}  \\
      Learning rate schedule~~ & cosine & cosine \\
      Weight decay & 0.03 & 0.03  \\
      $r$ in GR  & - & \{0.05, 0.1\} \\
      $\frac{\lambda}{r}$ in GR  & - & \{0.8, 1\} \\
      \bottomrule
    \end{tabular}
    \label{tbl : training hyperparameters vit}
  \end{table}

\section{Additional Results}
In additional to the classic ViT architecture family, we have also evaluated our methods with Swin and CaiT architecture families. Table \ref{tbl : other transformer warmup} shows the corresponds results on Cifar dataset. It should be mentioned that the common training hyper-parameters are the same as those of training ViT models.

As evidenced by the Table \ref{tbl : other transformer warmup}, our methods demonstrate improved performance across different Transformer architectures, which means that warmup GR can be essential for Transformer-based architecture. In fact, despite these models featuring distinct structural elements, their core mechanisms remain consistent with the conventional Transformer. It should be noted that among the three GR warmup strategy, zero-warmup strategy yield the best results for the most cases.

Meanwhile, we have also extended our evaluation to include TinyImageNet as well as ImageNet, as shown at Table \ref{tbl : imagenet}. From the results, our methods shows the effectiveness, as evidenced by a consistent reduction in error rates across all models tested.

\begin{table}[http]
    \vskip -0.05in
    \caption{Test error rate of Swin and CaiT models using Adam optimizers on the Cifar-\{10, 100\} datasets when trained with our three proposed GR warmup strategies}
    \vskip -0.05in
    \begin{center}
    \begin{small}
    \begin{sc}
    \begin{tabular}{lcccc}
        \toprule
        Model & ~Base~ & ~~~GR Warmup~~~ & ~~~Cifar-10~~~ & ~~~Cifar-100~~~ \\
        Architecture~~~& Optimizer & Policy & Error[\%] & Error[\%] \\
        \midrule
        \multirow{5}{*}{Swin-Tiny} & \multirow{5}{*}{Adam} & No GR  & $\ \ 12.39_{ \pm 0.21}\ $ & $\ \ 42.36_{ \pm 0.42}\ $ \\
        &  & Vanilla GR  & $\ \ 11.34_{ \pm 0.14}\ $ & $\ \ 40.65_{ \pm 0.33}\ $ \\
        \cmidrule{3-5}
        &  & $r$-Warmup  & $\ \ 10.54_{ \pm 0.19}\ $ & $\ \ 39.44_{ \pm 0.29}\ $ \\
        &  & $\lambda$-Warmup  & $\ \ 10.70_{ \pm 0.29}\ $ & $\ \ \mathbf{39.13_{ \pm 0.36}}\ $ \\
        &  & Zero-Warmup  & $\ \ \mathbf{10.52_{ \pm 0.16}}\ $ & $\ \ 39.51_{ \pm 0.32}\ $ \\
        \midrule
        \multirow{5}{*}{Swin-Small} & \multirow{5}{*}{Adam} & No GR  & $\ \ 12.72_{ \pm 0.17}\ $ & $\ \ 43.14_{ \pm 0.44}\ $ \\
        &  & Vanilla GR  & $\ \ 11.85_{ \pm 0.13}\ $ & $\ \ 42.20_{ \pm 0.42}\ $ \\
        \cmidrule{3-5}
        &  & $r$-Warmup  & $\ \ 11.68_{ \pm 0.22}\ $ & $\ \ 40.82_{ \pm 0.34}\ $ \\
        &  & $\lambda$-Warmup  & $\ \ 11.69_{ \pm 0.19}\ $ & $\ \ {40.69_{ \pm 0.40}}\ $ \\
        &  & Zero-Warmup  & $\ \ \mathbf{11.24_{ \pm 0.17}}\ $ & $\ \ \mathbf{40.12_{ \pm 0.26}}\ $ \\
        \midrule
        \multirow{5}{*}{Swin-Base} & \multirow{5}{*}{Adam} & No GR  & $\ \ 12.26_{ \pm 0.16}\ $ & $\ \ 41.63_{ \pm 0.53}\ $ \\
        &  & Vanilla GR  & $\ \ 11.88_{ \pm 0.20}\ $ & $\ \ 38.89_{ \pm 0.44}\ $ \\
        \cmidrule{3-5}
        &  & $r$-Warmup  & $\ \ 11.44_{ \pm 0.17}\ $ & $\ \ \mathbf{39.53_{ \pm 0.37}}\ $ \\
        &  & $\lambda$-Warmup  & $\ \ 11.34_{ \pm 0.27}\ $ & $\ \ {38.92_{ \pm 0.30}}\ $ \\
        &  & Zero-Warmup  & $\ \ \mathbf{10.90_{ \pm 0.15}}\ $ & $\ \ 38.69_{ \pm 0.42}\ $ \\
        \midrule
        \multirow{5}{*}{CaiT-XXS12} & \multirow{5}{*}{Adam} & No GR  & $\ \ 13.25_{ \pm 0.14}\ $ & $\ \ 41.74_{ \pm 0.46}\ $ \\
        &  & Vanilla GR  & $\ \ 13.62_{ \pm 0.23}\ $ & $\ \ 39.14_{ \pm 0.17}\ $ \\
        \cmidrule{3-5}
        &  & $r$-Warmup  & $\ \ 11.31_{ \pm 0.24}\ $ & $\ \ 38.74_{ \pm 0.33}\ $ \\
        &  & $\lambda$-Warmup  & $\ \ 12.06_{ \pm 0.31}\ $ & $\ \ \mathbf{39.31_{ \pm 0.43}}\ $ \\
        &  & Zero-Warmup  & $\ \ \mathbf{11.07_{ \pm 0.18}}\ $ & $\ \ 38.51_{ \pm 0.36}\ $ \\
        \midrule
        \multirow{5}{*}{CaiT-XXS24} & \multirow{5}{*}{Adam} & No GR  & $\ \ 11.89_{ \pm 0.19}\ $ & $\ \ 39.42_{ \pm 0.53}\ $ \\
        &  & Vanilla GR  & $\ \ 12.77_{ \pm 0.15}\ $ & $\ \ 38.02_{ \pm 0.36}\ $ \\
        \cmidrule{3-5}
        &  & $r$-Warmup  & $\ \ 10.47_{ \pm 0.13}\ $ & $\ \ \mathbf{37.04_{ \pm 0.41}}\ $ \\
        &  & $\lambda$-Warmup  & $\ \ 11.77_{ \pm 0.24}\ $ & $\ \ {37.52_{ \pm 0.29}}\ $ \\
        &  & Zero-Warmup  & $\ \ \mathbf{9.87_{ \pm 0.20}}\ $ & $\ \ 37.17_{ \pm 0.41}\ $ \\
        \midrule
        \multirow{5}{*}{CaiT-XS12} & \multirow{5}{*}{Adam} & No GR  & $\ \ 11.06_{ \pm 0.16}\ $ & $\ \ 39.15_{ \pm 0.47}\ $ \\
        &  & Vanilla GR  & $\ \ 12.52_{ \pm 0.19}\ $ & $\ \ 37.65_{ \pm 0.32}\ $ \\
        \cmidrule{3-5}
        &  & $r$-Warmup  & $\ \ \mathbf{9.48_{ \pm 0.24}}\ $ & $\ \ 37.83_{ \pm 0.36}\ $ \\
        &  & $\lambda$-Warmup  & $\ \ 10.23_{ \pm 0.21}\ $ & $\ \ {37.78_{ \pm 0.33}}\ $ \\
        &  & Zero-Warmup  & $\ \ {9.59_{ \pm 0.25}}\ $ & $\ \ \mathbf{37.46_{ \pm 0.49}}\ $ \\
        \midrule
        \multirow{5}{*}{CaiT-S12} & \multirow{5}{*}{Adam} & No GR  & $\ \ 10.59_{ \pm 0.18}\ $ & $\ \ 39.15_{ \pm 0.45}\ $ \\
        &  & Vanilla GR  & $\ \ 12.34_{ \pm 0.25}\ $ & $\ \ 36.41_{ \pm 0.41}\ $ \\
        \cmidrule{3-5}
        &  & $r$-Warmup  & $\ \ 9.03_{ \pm 0.29}\ $ & $\ \ 35.70_{ \pm 0.33}\ $ \\
        &  & $\lambda$-Warmup  & $\ \ 9.20_{ \pm 0.31}\ $ & $\ \ {36.12_{ \pm 0.43}}\ $ \\
        &  & Zero-Warmup  & $\ \ \mathbf{8.94_{ \pm 0.14}}\ $ & $\ \ \mathbf{35.16_{ \pm 0.39}}\ $ \\
        \bottomrule
    \end{tabular}
    \label{tbl : other transformer warmup}
\end{sc}
\end{small}
\end{center}
\end{table}

\begin{table*}[t]
    \vskip -0.05in
    \caption{Test error rate of ViT models using Adam optimizers on the TinyImageNet and ImageNet datasets when trained with our three proposed GR warmup strategies}
    \vskip -0.05in
    \begin{center}
    \begin{small}
    \begin{sc}
    \begin{tabular}{lccc}
        \toprule
        Model & ~Base~ & ~~~GR Warmup~~~ & ~~~Testing~~~ \\
        Architecture~~~& Optimizer & Policy & Error[\%] \\
        \midrule
        \multirow{5}{*}{ViT-TI} & \multirow{5}{*}{TinyImageNet} & No GR  & $\ \ 52.15_{ \pm 0.16}\ $ \\
        &  & Vanilla GR  & $\ \ 51.08_{ \pm 0.19}\ $ \\
        \cmidrule{3-4}
        &  & $r$-Warmup  & $\ \ 50.41_{ \pm 0.21}\ $ \\
        &  & $\lambda$-Warmup  & $\ \ 50.54_{ \pm 0.14}\ $ \\
        &  & Zero-Warmup  & $\ \ \mathbf{50.19_{ \pm 0.20}}\ $ \\
        \midrule
        \multirow{5}{*}{ViT-S} & \multirow{5}{*}{TinyImageNet} & No GR  & $\ \ 52.28_{ \pm 0.23}\ $ \\
        &  & Vanilla GR  & $\ \ 50.64_{ \pm 0.18}\ $ \\
        \cmidrule{3-4}
        &  & $r$-Warmup  & $\ \ 50.07_{ \pm 0.24}\ $ \\
        &  & $\lambda$-Warmup  & $\ \ 50.25_{ \pm 0.26}\ $ \\
        &  & Zero-Warmup  & $\ \ \mathbf{49.91_{ \pm 0.19}}\ $ \\
        \midrule
        \multirow{5}{*}{ViT-B} & \multirow{5}{*}{TinyImageNet} & No GR  & $\ \ 62.23_{ \pm 0.21}\ $ \\
        &  & Vanilla GR  & $\ \ 49.96_{ \pm 0.23}\ $ \\
        \cmidrule{3-4}
        &  & $r$-Warmup  & $\ \ 49.30_{ \pm 0.14}\ $ \\
        &  & $\lambda$-Warmup  & $\ \ 48.73_{ \pm 0.15}\ $ \\
        &  & Zero-Warmup  & $\ \ \mathbf{48.75_{ \pm 0.20}}\ $ \\
        \midrule
        \multirow{5}{*}{Swin-Tiny} & \multirow{5}{*}{ImageNet} & No GR  & $\ \ 23.43_{ \pm 0.24}\ $ \\
        &  & Vanilla GR  & $\ \ 22.34_{ \pm 0.15}\ $ \\
        \cmidrule{3-4}
        &  & $r$-Warmup  & $\ \ 21.81_{ \pm 0.28}\ $ \\
        &  & $\lambda$-Warmup  & $\ \ 22.04_{ \pm 0.22}\ $ \\
        &  & Zero-Warmup  & $\ \ \mathbf{21.58_{ \pm 0.34}}\ $ \\
        \bottomrule
    \end{tabular}
    \label{tbl : imagenet}
\end{sc}
\end{small}
\end{center}
\end{table*}

\end{document}